\documentclass{sig-alternate}

\begin{document}
\clubpenalty=10000
\widowpenalty = 10000

\title{Feature Selection Based on Term Frequency and T-Test for Text Categorization}
%
%
%
%
%

\numberofauthors{4} 
%
\author{
%
%
\alignauthor
Deqing Wang\\
       \affaddr{SKLSDE, Beihang University}\\
       \affaddr{Beijing, 100191, China}\\
       \email{dqwang@nlsde.buaa.edu.cn}
\and
\alignauthor
Hui Zhang\\
      \affaddr{SKLSDE, Beihang University }\\
       \affaddr{Beijing, 100191, China}\\
       \email{hzhang@nlsde.buaa.edu.cn}
\and
\alignauthor
Rui Liu, Weifeng Lv\\
      \affaddr{SKLSDE, Beihang University }\\
       \affaddr{Beijing, 100191, China}\\
       \email{\{liurui,lwf\}@nlsde.buaa.edu.cn}
}



\maketitle

\begin{abstract}
Much work has been done on feature selection. Existing methods are based on document frequency, such as Chi-Square Statistic, Information Gain etc. However, these methods have two shortcomings: one is that they are not reliable for low-frequency terms, and the other is that they only count whether one term occurs in a document and ignore the term frequency. Actually, high-frequency terms within a specific category are often regards as discriminators.

This paper focuses on how to construct the feature selection function based on term frequency, and proposes a new approach based on $t$-test, which is used to measure the diversity of the distributions of a term between the specific category and the entire corpus. Extensive comparative experiments on two text corpora using three classifiers show that our new approach is comparable to or or slightly better than the state-of-the-art feature selection methods (i.e., $\chi^2$, and IG) in terms of macro-$F_1$ and micro-$F_1$.
\end{abstract}




\section{Introduction}
Text classification (TC) is to assign new unlabeled natural language documents to  predefined thematic categories~\cite{Sebastiani02}. Many classification
algorithms have been proposed for TC, e.g., $k$-nearest neighbors~\cite{Yang97}, centroid-based classifier~\cite{Han00PKDD}, and support vector machines (SVMs)~\cite{CORTES95}.

Generally, text feature space is often sparse and high-dimensional. For instance, the dimensionality of a moderate-sized text corpus can reach up to tens or hundreds of thousands. The  high dimensionality of feature space will cause the ``curse of dimensionality'', increase the training time, and affect the accuracy of classifiers~\cite{Sebastiani02,Guyon03,Yang97}. Therefore, feature selection techniques are proposed to reduce the dimensionality under the premise of guaranteeing the performance of classifiers. Existing feature selection methods are based on statistical theory and information theory, such as $\chi^2$, IG, MI, and ECE. The theoretical basis of the four methods is sound, but the performances of these methods on TC tasks are different. Both $\chi^2$ and IG often achieved  better accuracy than MI and document frequency (DF)~\cite{Yang97}. However, other authors suspected the performance of IG on skewed text corpora~\cite{Mladenic99}.

Besides the classical methods, many improved methods have been proposed. For example, 
Yang et al.~\cite{YangSM02} considered the terms whose relative term frequency was larger  than a predefined threshold $\lambda$, and then modified the IG formula to select features.  Forman~\cite{Forman03} proposed the Bi-Normal Separation  (BNS) method, which used the standard Normal distribution's inverse cumulative probability function to construct feature selection function. 
 Uguz~\cite{Uguz11} proposed a two-stage feature selection method for TC by combining IG, principal component analysis and genetic algorithm. More and more methods have been generated, such as, mr2PSO~\cite{Unler11}, and improved TFIDF method~\cite{Wei08}. It is worth noting that $t$-test has been used for gene expression and genotype data ~\cite{Tibshirani02,zhou07}. However,  the variable in gene expression or genotype data is different from that in text data, i.e., the term frequency. Thus we try to validate the role of $t$-test in text feature selection.

From  document frequency perspective, the above methods almost use DF sufficiently. However,  no efficient method is proposed from term frequency perspective. It inspires our motivation of this paper. Our paper makes the following contributions:

(1) Using central limit theorem (CLT), we prove that the frequency distribution of a term  within a specific category or within the entire collection will be approximately normally distributed.

(2) We model the diversity of the frequency of a term between the specific category and the entire corpus with $t$-test. It means that if the distribution of one term within the specific category is obviously different with that within the entire corpus, the term can be considered to be feature.

(3) We verify our new approach on two common text corpora with three well-established classifiers. The experiments show that our approach is comparable to or even slightly better  than the state-of-the-art $\chi^2$ and ECE in terms of both macro-$F_1$ and micro-$F_1$ , and it outperforms IG and MI methods significantly on unbalanced text corpus.


\section{Feature Selection Metrics}\label{sec:fs}

 Many feature selection approaches have been proposed in TC tasks, but  we only give detailed analysis on four methods because they have been widely used and achieved better performance, the formulae can be found in Refs ~\cite{Yang97,Forman03,Guyon03}. They are: Chi-Square Statistic ($\chi^2$), Information Gain (IG), Mutual Information (MI), and Expected Cross-Entropy (ECE).
%


$\chi^2$ was proposed by Pearson early in 1900~\cite{Yang97}. The $\chi^2$ statistic is used to measure the lack of independence between $t_i$ and $C_j$, and can be regards as the $\chi^2$ distribution with one degree of freedom. In real-world corpus, $\chi^2$ statistic is based, however, on several assumptions that do not hold for most textual analysis~\cite{Dunning93}. For instance,  if term $t_1$ occurs in 50\% documents of a specific category $C_j$ and term  $t_2$ occurs in 49\% documents, but the frequency of $t_2$ is much higher than that of $t_1$. Experts often think term $t_2$ should have more discriminating power than $t_1$ in the specific category $C_j$. $\chi^2$, however, will be prone to select term  $t_1$ as feature, rather than $t_2$. The problem is that $\chi^2$ is not reliable for low-frequency terms~\cite{Dunning93}. 





The weakness of MI is that the score is strongly influenced by the marginal probabilities of
terms, because rare terms will have a higher score than common terms. Therefore,  the scores  are not comparable across terms of widely differing frequency~\cite{Yang97,Li09}. Besides, MI gives longer documents higher weights in the estimation of the feature scores. 



IG was firstly used as  attribute selection measure  in decision tree~\cite{Yang97}. This measure is from entropy in information theory, which studies the value or ``information content'' of messages. IG is defined as the difference between the original information requirement (i.e., based on just the proportion of classes) and the new requirement (i.e., obtained after partitioning on term $t_i$). IG is also called average mutual information. The weakness of IG method is that  it prefers to select terms distributed in many categories, but these terms have less discriminating power in TC tasks.
Differing from IG, Expected Cross-Entropy (ECE)~\cite{Koller97} only considers the  terms occurred in a document and ignores the absent terms.




 As we know, if a term (except stop words) occurs frequently within a specific category, the term should be considered as a feature or discriminator of the category. For example,  ``computer" occurs frequently in the IT category.   However, the above methods are all based on document frequency, and {\bf ignore} the term frequency. In next section, we will propose a new approach based on term frequency, and it can capture the information of high-frequency terms.

\section{New Approach Based on term frequency and t-test}\label{newmethod}

The $t$-test, namely the student $t$-test, is often used to assess whether the means
of two classes are statistically different from each other by calculating a ratio between the difference of two class means and the variability of the two classes~\cite{zhou07}. In this section, we explain why the averaged term frequency within a single category or in the whole corpus is approximately normal using Lindeberg-Levy central limit theorems, and then how the $t$-test is constructed based on the averaged term frequencies.

Let us consider the term frequency in  text corpus consisting of $n$ documents. Given a vocabulary V, the term frequency  ($tf_{ij}$) of a term $t_i~(1 \leq i \leq |V|)$ in the $j$th document $(1 \leq j \leq N)$ can be considered as a random variable, which  subjects to some unknown distribution, e.g.,  multinomial model~\cite{McCallum98}. In the multinomial model, a document is an ordered sequence of word events drawn from the same vocabulary V, and the probability of each word event in a document is independent of the word's context and position in the document. Therefore, each document $d_j$ is drawn from a multinomial distribution of words with as many independent trials~\cite{McCallum98}. That is, the occurrence of one term in each document is dominated by a multinomial function. Then,

 (1) Let $\{tf_{i1},\cdots,tf_{iN}\}$  be a random sample of size $N$, where $N$ is the number of documents in the collection, and $tf_{ij}(0 \leq j \leq N)$ is the term frequency of $t_i$ in $j$th document.  That is, a sequence of independent and identically distributed random variables with expected values $\mu_i=Np_i$ and variances $\sigma_{i}^2=Np_i(1-p_i)$, where $p_i$ is the distributed probability of term $t_i$ in the collection. Each sample belongs to one of $K$ classes $1, 2, \cdots, K$.

(2) Let $\overline{tf_i}=\frac{1}{N}(tf_{i1}+tf_{i2}+\cdots+tf_{iN})$ be the sample average of these random variables in terms of $t_i$.

(3) Let $\overline{tf_{ki}}= \sum_{j=1}^N tf_{ij}I(d_j,C_k)/N_k, (k=1,\cdots,K)$ be the sample average of term $t_i$ in category $C_k$, where $I(d_j,C_k)$ is an indicator to discriminate whether document $d_j$ belongs to $C_k$, and  $N_k$ is the total samples in class $k$.

 According to Lindeberg-Levy central limit theorems (LV CLT)~\cite{Billingsley95}, $\overline{tf_i}$ is  approximately normal with mean $\mu_i$ and variance $\frac{1}{N}\sigma_{i}^2$, denoted as $\tilde{N}(\mu_i, \frac{1}{N}\sigma_{i}^2)$;  And $\overline{tf_{ki}}$ is approximately normal with mean $\mu_i$ and variance $\frac{1}{N_k}\sigma_{i}^2$, denoted as $\tilde{N}(\mu_i, \frac{1}{N_k}\sigma_{i}^2)$.

Then we know that $\overline{tf_{ki}}-\overline{tf_i}$ is also approximately normal distributed with mean $0$ and variance $(\frac{1}{N_k}-\frac{1}{N})\sigma_{i}^2$. The variance (Var) is induced as follows:
\begin{eqnarray}\label{equ:d}
&&Var(\overline{tf_{ki}}-\overline{tf_i})\nonumber\\
&&=Var\big((\frac{1}{N_k}-\frac{1}{N})\sum_{j\in C_k}tf_{ij}+\frac{1}{N}\sum_{j \notin C_k}tf_{ij}\big)\nonumber\\
&&=\frac{(N-N_k)^2\times N_k\times \sigma_{i}^2}{N^2\times {N_k}^2}+\frac{(N-N_k)\times \sigma_{i}^2}{N^2} \nonumber\\
&&=(\frac{1}{N_k}-\frac{1}{N})\times \sigma_{i}^2.
\end{eqnarray}

Besides, we define the pooled within-class deviation as follows:
\begin{equation}\label{equ:deviation}
{s_i}^2 = \frac{1}{N-K}\sum_{k=1}^K\sum_{j\in C_k}(tf_{ij}-\overline{tf_{ki}})^2
\end{equation}

According to the definition of the $t$-test~\cite{William1908}, we construct the following formula:
 \begin{equation}\label{equ:ttest}
t-test(t_i,C_k)=\frac{\left\vert\overline{tf_{ki}}-\overline{tf_i}\right\vert}{m_k\cdot s_i}
\end{equation}
\noindent where $s_i$ is standard deviation, and $m_k=\sqrt{\frac{1}{N_k}-\frac{1}{N}}$.

The Eq.~\ref{equ:ttest} is used to measure whether the means of the two normal distributions (i.e., $\overline{tf_{ki}}$ and $\overline{tf_i}$) have the  statistically significant difference. The bigger the value of $t-test(t_i,C_k)$ is, the larger the difference of the means  is. For some threshold $\theta$, if the $t-test(t_i,C_k)< \theta$ , it implies that the averaged frequency of term $t_i$ in the specific category $C_k$ has the same or similar mean  with that in the entire corpus; Otherwise, it implies  the averaged frequency of term $t_i$ in the specific category $C_k$ is significantly different from  that in the entire corpus, and the term has more discriminating power for the specific category $C_k$.  Compared with the average of term frequency in the entire corpus, the term $t_i$ occurred many or few times in $C_k$ can be considered as the feature of category $C_k$.

We combine the category-specific scores of a term into two alternate ways:
\begin{equation}\label{equ:sum}
t-test_{avg}(t_i)=\sum_{k=1}^K t-test(t_i,C_k)
\end{equation}

\begin{equation}\label{equ:tmax}
t-test_{max}(t_i)=\max_{k=1}^K\{ t-test(t_i,C_k)\}
\end{equation}

\section{Experimental Setup}\label{setup}

\subsection{Data Sets}

\textit{Reuters-21578}~\footnote{Available on http://ronaldo.cs.tcd.ie/esslli07/sw/step01.tgz}: The Reuters corpus is a
widely used benchmark collection~\cite{Dunning93,Forman03,Yang97,YangSM02}. According to the ModApte split, we get a
collection of 52 categories (9100 documents) after removing unlabeled documents and
documents with more than one class label. Reuters-21578 is a very skewed data set. Altogether 319 stop words, punctuation and numbers areremoved. All letters are converted into lowercase, and the word
stemming is  applied.

\textit{20Newsgroup}~\footnote{Available on http://kdd.ics.uci.edu/databases/20newsgroup.}: The Newsgroup is also a widely used benchmark~\cite{Dunning93,Forman03,Yang97}, and
consists of 19,905 documents, which are uniformly distributed in twenty
categories. We randomly divide it into training and test sets by 2:1, and only keep ``Subject'', ``Keyword'' and ``Content''.
The stop words list has 823 words, and we filter words containing non-characters. All letters are
converted into lowercase and word stemming is applied.

Each document is represented by a vector in the term space, and term weighting is calculated by standard $ltc$~\cite{Salton88}, and then the vector is normalized to have one unit length.

\subsection{Classifiers}

In our experiments, we choose three well-established classifiers for the comparison purpose. They are: Support Vector Machines (SVMs)~\cite{CORTES95}, weighted kNN classifier ($k$NN)~\cite{Yang97}, and classic Centroid-based Classifier (CC)~\cite{Han00PKDD}. The SVMs implementation we use is LIBSVM~\cite{Chang01} with linear
kernels. For $k$NN, we set $k=10$ ~\cite{Yang97}. The similarity measure we use
is the cosine function.

\subsection{ Performance Measures}
We measure the effectiveness of classifiers
in terms of $F_1$ widely used for TC.
For multi-class task, $F_1$ is estimated in two ways, i.e., the macro-averaged $F_1$ (macro-$F_1$) and the micro-averaged $F_1$ (micro-$F_1$), as the following:
\begin{equation}
\textrm{macro-}F_1=\frac{\sum_{i=1}^K F_1(i)}{K},
\end{equation}
\begin{equation}
\textrm{micro-}F_1=\frac{2\bar{p}\bar{r}}{\bar{p}+\bar{r}},
\end{equation}
\noindent where $F_1(i)$ is the $F_1$ value of the predicted $i$th class, and $\bar{p}$ and $\bar{r}$ are the precision and recall values across all classes, respectively.
In general, macro-$F_1$ gives the same weight to all categories. In contrast, micro-$F_1$ gives the same weight to each instance, which can be dominated by the performance of common or majority categories.

\section{Results}\label{results}
\setcounter{subsection}{0}
Firstly, We show one case study of $t$-test in real-world corpus. Tables~\ref{tab:keywordsexample} lists the scores of
seven different feature selection functions for the selected four terms in
category ``acq'' from the real-life corpus, i.e., Reuters-21578. Based on the literal meaning, the first two
terms, i.e., ``acquir" and ``stake", are closely related to the content
of category ``acq'', while the last two terms, i.e., ``payout" and
``dividend", belong to other category. However, according to the $\chi^2$ , ECE, and TF methods, we wrongly select ``acquir" and ``dividend" as the features of category ``acq'', whereas $t$-test, IG and MI select the features correctly.

\begin{table}[thb]
\centering
\caption{The feature values of four terms in  ``acq".}
 \begin{tabular}{ccccc}
  \hline
          &acquir&stake&payout&dividend\\
  \hline
  $t-test$&28.053&22.567&3.272&17.796\\
  $\chi^2$&479.482&270.484&131.104&344.045\\
  $IG$&0.078&0.042&0.009&0.036\\
  $MI$&1.283&1.126&0.362&0.830\\
  $ECE$&0.084&0.050&0.028&0.060\\
  $TF$&749&646&232&903\\

  \hline
 \end{tabular}
 \label{tab:keywordsexample}
\end{table}

Then, we show the performance of $t$-test on two corpora with three classifiers. For Reuters-21578, the number of feature space is all, 17000,  15000,  13000, 11000, 10000, 8000, 6000, 4000, and 2000, respectively, accounting to ten groups of data sets. On 20 Newsgroup corpus, the original feature space reaches up to 210 thousand and we only select less terms as features to save training time. The dimensionality of feature space is all, 2000, 1500, 1000, 500, and 200, respectively, accounting to  six groups of data sets.

For $\chi^2$, MI, and $t$-test methods, we tested the two alternative combinations, i.e., $averaged$ and $maximized$ ways. We observed that the averaged way was always better than the maximized way for multi-classes problem. Thus we only report the best results of three methods.

\subsection{Performance of t-test with kNN classifier}
\begin{figure}[thb]
  \centering
  \includegraphics[width=0.5\textwidth]{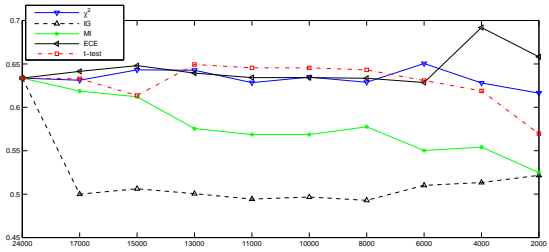}
  \caption{The comparative curves of five  methods with $k$NN on Reuters-21578  in terms of macro-$F_1$.}
  \label{fig:maKNNRe}
\end{figure}

\begin{figure}[thb]
  \centering
  \includegraphics[width=0.5\textwidth]{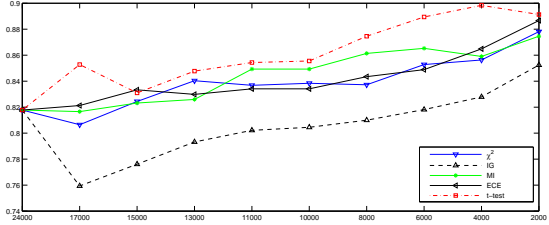}
  \caption{The comparative curves of five  methods with $k$NN on  Reuters-21578 in terms of micro-$F_1$.}
  \label{fig:miKNNRe}
\end{figure}

The macro-$F_1$ and micro-$F_1$ of five  methods with $k$NN  on imbalanced Reuters-21578 are shown in Fig.~\ref{fig:maKNNRe}, Fig.~\ref{fig:miKNNRe}, respectively. It is clear that $t$-test, $\chi^2$, and ECE achieve evidently better performance than MI and IG in terms of macro-$F_1$. However, the diversity among the three methods is small. As shown in Fig.~\ref{fig:maKNNRe}, when the number of feature space is larger than 13000,  $\chi^2$, and ECE is a little better than  $t$-test; However, when the number of features falls in [8000, 13000], $t$-test performs the best macro-$F_1$.

The micro-$F_1$ of five  methods increases  as the number of features  decreases, as shown in Fig.~\ref{fig:miKNNRe}. It demonstrates that  $k$NN  often obtains better performance with less features. Our $t$-test method  performs consistently the best in distinct feature dimensionality, and the highest micro-$F_1$ of $t$-test is 89.8\% when the number of features is 4000, which improves up to 4.2\% than $\chi^2$. IG  achieves the worst performance in the all experiments on skewed corpus with $k$NN.

As shown in Fig.~\ref{fig:maKNNRe} and Fig.~\ref{fig:miKNNRe}, for unbalanced multi-class tasks, we find IG is inferior to MI in terms of both macro-$F_1$ and micro-$F_1$, whereas IG is superior to MI for binary classification tasks according to the comparative experiments of Yang et al~\cite{Yang97}. The conflict shows that feature selection methods depends on the practical classification problem.

\begin{figure}[thb]
  \centering
  \includegraphics[width=0.5\textwidth]{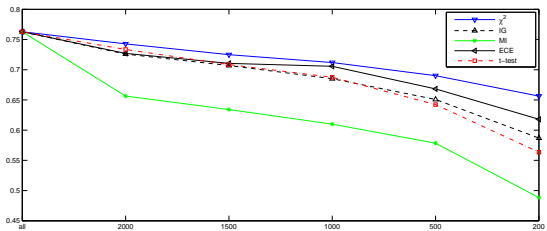}
  \caption{The comparative curves of five  methods with $k$NN on 20 Newsgroup  in terms of micro-$F_1$.}
  \label{fig:miKNNNews}
\end{figure}

Because  macro-$F_1$ on balanced corpus is close to  micro-$F_1$ , we only show the results of micro-$F_1$ on 20 Newsgroup. As shown in Fig.~\ref{fig:miKNNNews}, the micro-$F_1$  of both $\chi^2$ and IG are slightly better than our $t$-test method, and the four methods are obviously better than MI. Especially, the performance of IG is comparable to $\chi^2$,  and ECE on balanced corpus.

\subsection{Performance of t-test with SVMs classifier}

\begin{figure}[thb]
  \centering
  \includegraphics[width=0.5\textwidth]{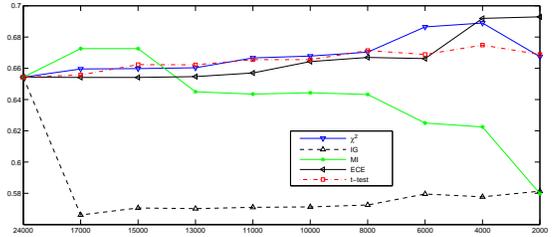}
  \caption{The macro-$F_1$ of different methods on Reuters-21578 using SVMs.}
  \label{fig:maSVMRe}
\end{figure}

\begin{figure}[thb]
  \centering
  \includegraphics[width=0.5\textwidth]{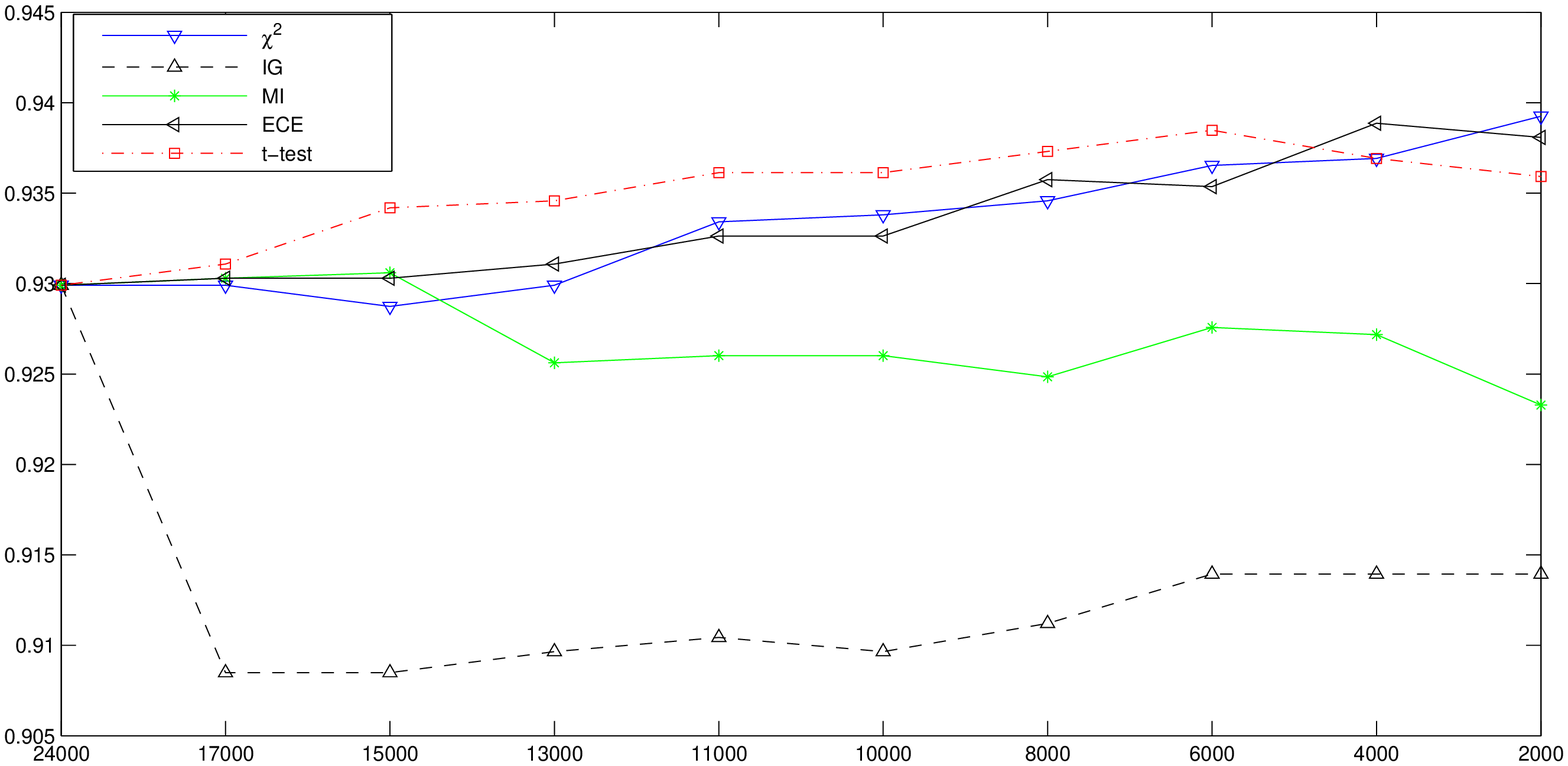}
  \caption{The micro-$F_1$ of different methods on Reuters-21578 using SVMs.}
  \label{fig:miSVMRe}
\end{figure}

Fig.~\ref{fig:maSVMRe} and Fig.~\ref{fig:miSVMRe} depict the macro-$F_1$ and micro-$F_1$  of different methods on the Reuters-21578 corpus using SVMs. The $t$-test, $\chi^2$,  and ECE  methods perform similar performances, which are better than IG and MI methods. Meanwhile, the macro-$F_1$ scores of three methods increase as the number of features reduces.  It is worth noting that MI does better than other methods when the number of features is in [15,000, 24,411], and then MI falls dramatically.

The performance of these methods in terms of micro-$F_1$  on Reuters-21578 corpus is shown in Fig.~\ref{fig:miSVMRe}. The micro-$F_1$ points of different feature selection methods show a tendency to increase as the number of the features decreases. However, these methods show consistent performance in micro-$F_1$, and the $t$-test method is still the best among these methods.

\begin{figure}[thb]
  \centering
  \includegraphics[width=0.5\textwidth]{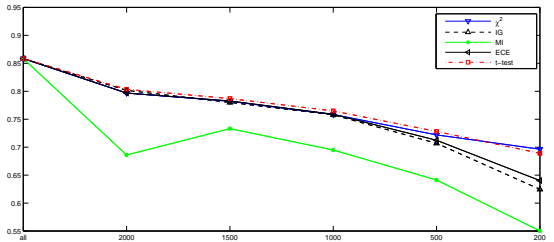}
  \caption{The micro-$F_1$ of different methods on 20 Newsgroup using SVMs.}
  \label{fig:miSVMNews}
\end{figure}

Fig.~\ref{fig:miSVMNews} depicts the micro-$F_1$ of different methods on the
20 Newsgroups  using SVM. The trends of the curves are similar to those in Fig.~\ref{fig:miKNNNews}. The $t$-test, $\chi^2$, IG, and ECE achieve similar performances, which are better than  MI. Our $t$-test is slightly better than others.

\subsection{Performance of t-test with Centroid-based classifier}

\begin{figure}[thb]
  \centering
  \includegraphics[width=0.5\textwidth]{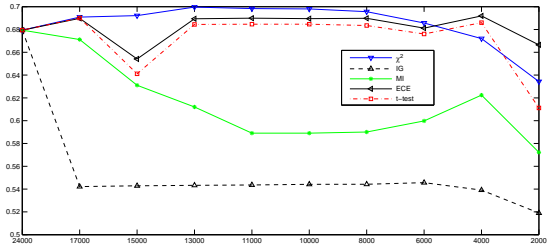}
  \caption{The macro-$F_1$ of five methods on Reuters-21578 using centroid-based classifier.}
  \label{fig:maCCRe}
\end{figure}

\begin{figure}[thb]
  \centering
  \includegraphics[width=0.5\textwidth]{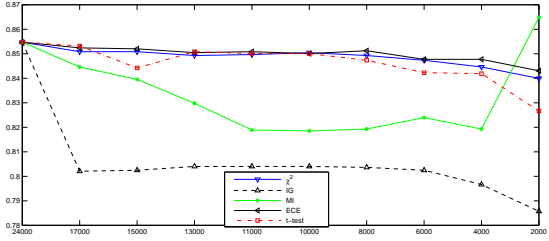}
  \caption{The micro-$F_1$ of five methods on Reuters-21578 using centroid-based classifier.}
  \label{fig:miCCRe}
\end{figure}

For centroid-based classifier, the macro-$F_1$ of five methods is shown in Fig.~\ref{fig:maCCRe}. We can observe that  $\chi^2$,  ECE,  and $t$-test do better than MI and IG methods, and $\chi^2$  is slightly better than   ECE and $t$-test. The same conclusion can be done in terms of micro-$F_1$, as shown in Fig.~\ref{fig:miCCRe}.

Meanwhile, our $t$-test is slightly better than $\chi^2$,  ECE, and IG methods on 20 Newsgroup corpus. The four methods outperform the MI method significantly.

\begin{figure}[thb]
  \centering
  \includegraphics[width=0.5\textwidth]{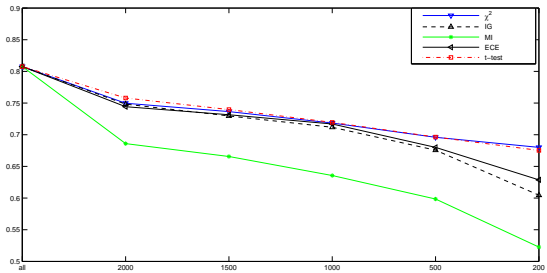}
  \caption{The micro-$F_1$ of five methods on 20 Newsgroup using centroid-based classifier.}
  \label{fig:miCCNews}
\end{figure}

\section{Conclusion and Future Work}\label{con}

In this paper, we  proposed a new feature selection method based on term frequency and $t$-test.  Then we compare our approach with the state-of-the-art methods on two corpora using three classifiers in terms of macro-$F_1$ and micro-$F_1$.  Extensive  experiments  have indicated that our new approach offers comparable performance with $\chi^2$, and ECE, even slightly better than them.  In future work, we will verify our method on more  text collections.

%


%
%
\balancecolumns

\end{document}